# EEG-Fest: Few-shot based Attention Network for Driver's Drowsiness Estimation with EEG Signals

Ning Ding, Ce Zhang and Azim Eskandarian

Mechanical Engineering, Virginia Polytechnic Institute and State University, Blacksburg, VA, USA

E-mail: ningding@vt.edu, zce@vt.edu, and eskandarian@vt.edu

**Abstract**

The leading factor behind most vehicular accidents is the driver's inattentiveness. To accurately determine driver's drowsiness, Electroencephalography (EEG) has been proven to be a reliable and effective method. Even though previous studies have developed accurate driver's drowsiness detection algorithms, certain obstacles still persist, such as (a) limited training sample sizes, (b) detecting anomalous signals, and (c) achieving subject-independent classification. In this paper we propose a novel solution, names as EEG-Fest, which is a generalized few-shot model aimed at addressing the aforementioned limitations. The EEG-Fest has the ability to (a) classify a query sample's level of drowsiness with only a few support sample inputs (b) identify whether a query sample is anomalous signals or not, and (c) perform subject-independent classification. During the evaluation, our proposed EEG-Fest algorithm demonstrates better performance compared to other two conventional EEG algorithms in cross-subject validation.

Keywords: Electroencephalography, Deep Learning, Few-shot Learning, Attention Neural Network

## 1. Introduction

The National Highway Traffic Safety Administration (NHTSA) crash report indicates that more than 90,000 accidents occur as a result of driver drowsiness [1, 2]. Despite the rapid progress in autonomous vehicle technology, current solutions still require human drivers to play a role in autonomous driving, such as keeping an eye on the surrounding environment and being ready to regain control of the vehicle during specific tasks [3,4]. Therefore, keeping track of driver drowsiness is essential for ensuring safe driving.

Currently, driver drowsiness estimation can be classified into two categories: (a) indirect measurement, which uses the vehicle's movement and performance to assess the driver's level of drowsiness [5,6], and (b) direct measurement, which utilizes physiological sensors to directly measure the driver's cognitive state [7-14]. While indirect measurement-based approaches are relatively straightforward, their reliability and performance for driver drowsiness estimation are often inferior to those of direct measurement-based approaches.

One of the most widely used direct measurement tools for analyzing driver drowsiness is Electroencephalography (EEG), a physiological sensor that measures brain signals in humans [15, 16]. Advanced pattern recognition algorithms are necessary to accurately decipher EEG signals, as these signals are frequently non-stationary and prone to contamination from noise sources such as eye movements [17]. Currently, most EEG-based driver drowsiness estimation methods are categorized into conventional machine learning (ML) and deep learning (DL) approaches. The ML-based algorithms require either manual or automatic signal pre-processing and feature extraction techniques to obtain relevant information from raw EEG signals. Then, the processed features are used by classifiers to predict the driver's level of drowsiness [18, 19]. The DL-based algorithms use end-to-end neural network models for automatic signal denoising, feature extraction, and classification [20]. According to recent research findings, DL-based approaches typically outperform ML-based approaches because of their larger number of parameters, higher-dimensional feature extraction, and effective neural network model architectures. In summary, existing algorithms can successfully estimate the driver drowsiness level with promising performance, and some algorithms can achieve real-time classification.

Despite these advancements, current algorithms still suffer from the following limitations: (a) many algorithms, particularly DL-based ones, require large amounts of training data to achieve satisfactory classification accuracy, (b) most algorithms are unable to identify anomalous signals such as noise or signals unrelated to driving drowsiness, and (c) current algorithms lack robustness and are often limited to subject-specific classification. These challenges hinder the development and practical application of EEG-based driver drowsiness detection systems, as collecting a large amount of data and conducting subject-specific experiments are both resource-intensive.

Thus, to address the shortcomings of existing algorithms, we present EEG-Fest, a subject-independent EEG-based few-shot network designed to both estimate driver drowsiness and detect anomalous signals. The main contributions and novelties of this study are shown as follows:

• We propose a new approach to estimating driver drowsiness and detecting anomalous signals through the use of EEG signals, utilizing few-shot learning. The performance of this model is assessed using data samples from a new subject that were not present during the training phase. To aid in the estimation of drowsiness and the detection of anomalous signals, only a small number of support samples are required from the new subject. To the best of our knowledge, this study should be the first work using few-shot learning method for both driver drowsiness estimation and anomalous signals detection.

• We incorporate both self-attention and cross-attention mechanisms in our proposed model. Our study indicates that the self-attention mechanism effectively extracts information from raw EEG signals more efficiently compared to previous methods, while the cross-attention mechanism highlights the connection between the support and query samples, providing a solid base for classifying the query samples. In our study, we show that attention mechanism is more suitable for detecting anomalous signals.

• Our proposed model is tested on various EEG datasets [21-23] to demonstrate its robustness and correctness.

The remaining sections of this paper is organized as follows: Section II reviews recent popular literature related to EEG-based driver drowsiness studies, few-shot learning, and attention-based neural networks. Section III elaborates on the details of our proposed method. Section IV and Section V report the results of our experiments and ablation study. Finally, Section VI offers the conclusion of our work.

## 2. Related Works

This section reviews the recent literature about EEG-based driver drowsiness estimation, few-shot learning algorithms, and attention-based neural networks.

### *2.1 EEG-based Driver Drowsiness Estimation*

Currently, there are two primary categories of EEG-based drowsiness detection algorithms: ML-based algorithms and end-to-end DL-based methods.

### *2.1.1 Conventional Machine Learning Approach*

Conventional ML methods for drowsiness detection typically follow a two-stage process. The first stage involves using explicit filtering techniques, such as bandpass filtering and time-frequency analysis, to extract EEG signals and perform feature extraction. The second stage involves using classifiers, such as support vector machine (SVM) or linear discriminant analysis (LDA), to estimate the drowsiness levels [24, 25]. The benefit of the conventional ML method is that it doesn't require a large-scale training process and its code implementation is relatively straightforward, making it a popular choice for researchers analysing driver EEG signals. W.L. Zheng, et al [22] constructed a Support Vector Regression model using a radial basis function kernel to investigate the relationship between driver drowsiness and EEG signals from different brain areas. However, there is no universally optimal feature or classifier algorithm for these tasks, and the appropriate choice of feature and classifier algorithms depends on the specific case. Jianfeng Hu [26] conducted a comparison of ten different classifiers and showed that the optimal combination of features and classifiers varies between subjects. Additionally, traditional manual signal processing and feature extraction are often unable to uncover implicit features generated by different subjects. As a result, many researchers are exploring ways to enhance the conventional ML methods, such as enhancing current signal pre-processing techniques [27], modifying existing classic classifiers, and proposing new EEG-based ML systems, to improve the accuracy of drowsiness estimation. Ruo-Nan Duan [28] suggested using a differential entropy feature for EEG analysis and demonstrated its superiority when compared to the conventional energy spectrum feature. Gang Li [29] enhanced the performance of support vector machine-based classification for detecting drowsiness in drivers by combining posterior classification probability and support vector machine classification together. Xue-Qin Huo [30] introduced distance information into an extreme learning machine to form a graph regularization term. The information from samples belonging to the same cluster can be taken into account in this term, thus restricting the output weights of the extreme learning machine algorithm. Dongrui Wu [31] merged fuzzy sets and domain adaptation to carry out online weighted adaptation regularization for regression problems, thereby obtaining improved estimation results. Despite these noteworthy advancements in research, prior studies have neglected to address anomalous signal detection, a common occurrence in real-world conditions.

### *2.1.2 Deep Learning Approach*

As neural networks and parallel computing techniques continue to advance, the use of deep neural networks has become prevalent for tackling high-dimensional classification and regression problems [32]. For example, Mehdi Hajinoroozi [33] employed a deep neural network on the sample covariance matrix computed from EEG epochs, resulting in a more accurate classification of drivers' drowsy states compared to shallow learning methods. With massive training samples, high-dimensional feature extractions, and non-linearity feature extraction algorithms, deep neural network-based methods tend to outperform traditional ML algorithms. Furthermore, despite the time-consuming training process, the processing speed of most neural networks during the forward operation is fast and can facilitate real-time driver drowsiness estimation tasks. Because of these benefits, many new deep-learning approaches have been proposed to tackle the Driving EEG task. Fu-Chang Lin's [34] Self-organizing Neural Fuzzy Inference Network (SONFIN) system was able to predict drivers' reaction time to an unexpected event by performing the Fuzzy rule in Neural Networks. Zhongke Gao [35] used convolutional layers and pooling layers to extract temporal information from EEG signals to evaluate drivers' fatigue. Nan Zhang [36] proposed two LSTM Neural Network architectures that accounted for the time-dependent nature of EEG signals, demonstrating improved performance for EEG-based vigilance estimation when compared to other non-temporal dependent models. Hong Zeng [37] developed a 5-layer convolution neural network to classify drivers' fatigue state by inputting raw EEG data into the model. The model outperforms the LSTM network. However, deep neural networks rely heavily on large amounts of data. Some researchers have found solutions for cases where limited data is available for training, such as Chun-Shu Wei [38], who constructed a model that leverages data from other subjects when subject-specific data is difficult to obtain in large quantities. Despite of this, we still hope that we can have an approach that can produce acceptable predictions with a limited number of samples from a particular subject.

*2.2 Few-Shot Learning*

Few-shot learning is a type of ML that trains a model to recognize new classes of subjects or states of a new subject using only a limited number of example samples and supervised information.[39] Like many other ML techniques, few-shot learning can handle both classification and regression problems. Few-shot classification determines the class of a query sample based on limited class samples in the support set. Our study belongs to this category of few-shot task. Few-shot regression estimates a regression function based on limited pairs of observed values and the dependent variable from an unknown function in the support set. [40] is an instance of a study in the field of few-shot regression. A successful few-shot classification model should be able to effectively differentiate between support sets and query sets. There are many types of few-shot models available for classification problems. Wei-Yu Chen [41] compared and analysed several well-known few-shot classification algorithms and studied the limitations of the standard evaluation framework. A neural network can also be adapted for use as a few-shot model. Jake Snell's [42] Prototypical Networks utilized a trained neural network to extract the feature space of each class in the support sets and then performed classification by using these feature spaces. Qianru Sun [43] employed a two-step approach where he first trained a deep neural network on large-scale data and then froze its parameters as a feature extractor, which helped it to rapidly adapt to few-shot tasks. Ruixiang Zhang [44] integrated generative adversarial networks (GANs) into few-shot learning, using fake samples generated by the GAN to train the model and sharpened the decision boundary between different classes. Inspired by prior successful efforts, our proposed approach utilizes a self-attention neural network module as the feature extractor, yielding exceptional results compared to other ML models. The feature extraction process for both support sets and query sets are followed by a cross-attention mechanism, which accentuates the relevance between the two sets and calculates their similarity, ultimately enhancing the classification performance.

*2.3 Attention Neural Network*

In our work, two attention neural networks are utilized: self-attention and cross-attention.

The self-attention mechanism has the ability to recall long sequences and focus on crucial elements in a new input sequence, resulting in a matrix that showcases the relevance of each element in the sequence. It has been demonstrated to be effective in language modeling and image recognition tasks. Adams Wei Yu [45] used self-attention mechanism to learn the global interaction between each pair of words. Hengshuang Zhao [46] showed that a model fully based on self-attention mechanism can outperform convolutional networks in image recognition tasks. Our work differs from prior works by focusing on EEG signals. We utilize the self-attention mechanism's ability to retain important information from a long sequence to derive the feature map of EEG within a particular time window, creating a solid foundation for further processing.

The cross-attention mechanism differs slightly from self-attention. Typically, the attention mechanism has three inputs. While all three inputs of the self-attention mechanism are identical, in the cross-attention mechanism, one of the three inputs is distinct from the other two. Hence, the cross-attention mechanism can produce a feature map that highlights the correlation between dissimilar inputs by preserving significant information from both. We use this benefit of the cross-attention mechanism to determine the similarity between support and query samples for classification. Its success in addressing few-shot image tasks [47] has been demonstrated, and our study shows its effectiveness for EEG tasks as well.

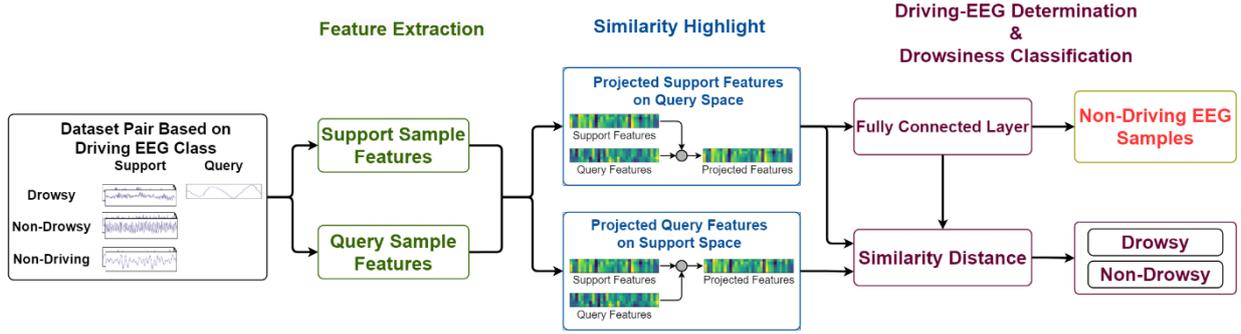

Figure 1: The EEG-Fest model comprises four main blocks: feature extraction, similarity highlighting, Driving-EEG determination, and drowsiness classification.

Table 1 EEG-Fest Terminology

| Terminology | Explanation | Acryname |
|---|---|---|
| driving-EEG | EEG collected during driving conditions. | / |
| non-driving EEG | EEG collected during non-driving conditions. | / |
| query sample | Sample whose class needs to be predited . | $x^q$ |
| support sample | Example sample for each class. | $x^s$ |
| N | The number of classes | / |
| K | The number of support samples for each class. | / |
| query set | Sample set including all query samples, where $n_q$ is the total number of query samples. | $Q = \{x_j^q\}_{j=1}^{n_q}$ |
| support set | Sample set containing N classes and K labeled samples per class, where $y^s$ is the ground truth of $x^s$, $n_s$ is the total number of support samples and ground truth, $n_s = N * K * n_q$. | $S = \{(x_n^s, y_n^s)\}_{n=1}^{n_s}$ |
| support subset | The subset of support set for the $i$-th class. | $S_i$ |
| paired data samples | Paired data samples consisting of the $j$-th query sample with support subset of each class. | $P_j = \{(S_i, x_j^q)\}_{i=1}^{N}$ |
| paired feature maps | The output of the feature extraction block for $P_j$, where $S_{fi}$ is the feature map got from $S_i$, $x_{fj}^q$ is the feature map got form $x_j^q$. | $P_{fj} = \{S_{fi}, x_{fj}^q\}_{i=1}^{N}$ |
| paired highlighted matrices | The output of the similarity highlighting block, where $S_{fi}^*$ is the matrix got from projecting $x_{fj}^q$ on $S_{fi}$, $x_{fj}^{q*}$ is the matrix got from projecting $S_{fi}$ on $x_{fj}^q$. | $P_{fj}^* = \{S_{fi}^*, x_{fj}^{q*}\}_{i=1}^{N}$ |

## 3. Methodology

The proposed EEG-Fest model is depicted in Figure 1, composed of four blocks: feature extraction, similarity highlighting, driving-EEG determination and drowsiness classification. Due to the model complexity, we illustrate main EEG-Fest terminologies in Table I.

Our model objective is to assess the driver drowsiness and determine whether the given EEG sample belongs to driving-EEG samples by comparing a query EEG sample with limited driving-EEG samples. The feature extraction block extracts feature maps from raw EEG sample, with the self-attention mechanism identifying the crucial information of the raw sample. The similarity highlighting block uses the cross-attention mechanism to accentuate the similarity between the feature maps of the support samples and the query sample. The driving-EEG determination and drowsiness classification blocks rely on the output from the similarity block to determine whether a query EEG sample belongs to driving-EEG samples and classify the driver's drowsiness.

In the subsequent paragraphs of this section, we provide an overview of EEG-Fest's few-shot learning basics, followed by an explanation of the feature extraction block, similarity highlighting block, driving-EEG determination block, and drowsiness classification block.

### 3.1 Few-Shot Learning Preliminaries

The goal of the few-shot learning approach is to address the scarcity of large-scale datasets caused by the significant expenses and human effort required for data collection. This approach is applicable to tasks that have only a few examples with supervised information [61]. In the few-shot approach, two sets of input are utilized: the support set and the query set. The support set is a collection of reference samples with ground truth labels, while the query set is needed to be predicted by being compared with the support set. Few-shot learning aims to perform estimation by identifying similarities and differences between the sample in the query set and a limited number of samples in the support set. The naming convention commonly used in few-shot learning is *N*-way *K*-shot learning, where *N* refers to the number of classes and *K* refers to the number of support samples.

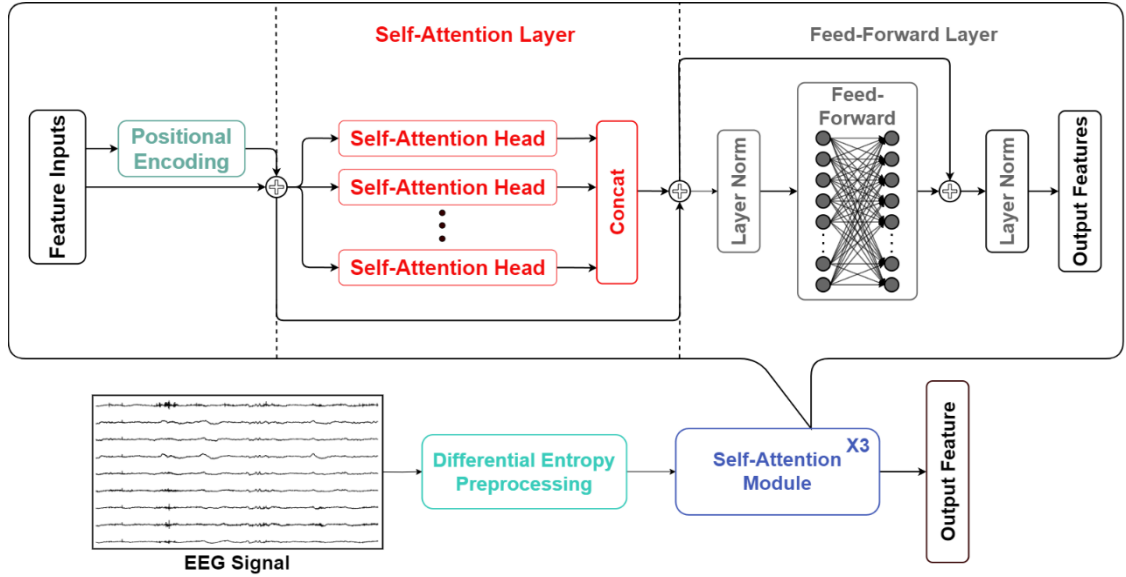

Figure 2: This figure is the overview of the feature extraction block. Initially, the differential entropy feature is extracted from the raw EEG data. Subsequently, a self-attention mechanism is employed to obtain the feature map, which selectively attends to the vital aspects of the differential entropy feature

In this study, the query set can be written as $Q = \{x_j^q\}_{j=1}^{n_q}$. The support set can be expressed as $S = \{(x_n^s, y_n^s)\}_{n=1}^{n_s}$. We denote $S_i$ as the support subset of the *i*-th class. By pairing the *j*-th query sample with support subset of each class, we obtain paired data samples $P_j = \{(S_i, x_j^q)\}_{i=1}^{N}$ $(j = 1, \cdots, n_q)$ for each query sample in query set. $P_j$ servers as the input of the proposed EEG-Fest.

### 3.2 Feature Extraction Block

The primary goal of the feature extraction block is to convert the raw EEG data into numerical features, thereby expanding the dimension of the data, and allowing for analysis from various perspectives. As output, the block generates feature maps $P_{fj} = \{S_{fi}, x_{fj}^q\}_{i=1}^{N}$ based on paired data samples $P_j$. As shown in Figure 2, the block comprises of a differential entropy (DE) preprocessing step for the raw EEG data and three self-attention modules. The DE preprocessing step transforms the raw data into DE features across five distinct frequencies. The self-attention modules are designed to extract and capture the most crucial information from the DE features.

### 3.2.1 Differential Entropy Preprocessing

Under the assumption that EEG samples conform to a Gaussian distribution, it is possible to compute the DE of each channel. Previous research [28] has demonstrated the success of DE as a technique for EEG-based classification, while additional studies [48, 49] have validated its effectiveness in EEG tasks. Therefore, we have decided to use the same DE technique for EEG preprocessing.

In EEG-Fest, the DE preprocessing step involves transforming the raw EEG data into five distinct frequency bands for each channel: delta (1-3Hz), theta (4-7Hz), alpha (8-13Hz), beta (14-30Hz), and gamma (31-50Hz). From each of these frequency bands, DE features are extracted for each channel of the EEG data. The equation to compute DE is:

$$f(x) = -\int_{-\infty}^{\infty} \frac{1}{\sqrt{2\pi\sigma^2}} e^{-\frac{(x-\mu)^2}{2\sigma^2}} \log \frac{1}{\sqrt{2\pi\sigma^2}} e^{-\frac{(x-\mu)^2}{2\sigma^2}} dx$$
$$= \frac{1}{2} \log(2\pi e \sigma^2) \quad (1)$$

where *x* is the EEG sample following Gaussian distribution $N(\mu, \sigma^2)$.

The DE for each channel across the five frequency bands can be arranged in a matrix *De* with a dimension of $5 \times c$, where c represents the number of channels in the EEG sample. After acquiring the matrix *De* from the raw EEG data, a linear transformation is applied to it to expand the dimension of the matrix, which gives rise to the matrix *D*.

$$D = De \cdot W + b \quad (2)$$

where, $W$ is learnable weight, $b$ is learnable bias. The output dimension of $D$ in this study is $5 \times 32$.

### 3.2.2 Self-Attention Module

The self-attention module is inspired by [50, 51], which has demonstrated impressive performance in both natural language and computer vision tasks. The self-attention module creates a correlation across the entire input sequence during the learning process and identifies significant information from these correlations. This characteristic is leveraged in our study, where we use the self-attention module with extended DE features of all EEG channels as input. This approach enables the model to learn and extract features from a global perspective. The self-attention module comprises three components: positional encoding, self-attention layer, and feed-forward layer. In this study, we use a sequence of three self-attention modules.

The positional encoding ($PE$) technique is based on the methodology proposed in [50]. The $i$-th position of the PE in sequence can be described as follows:

$$PE(p, i) = \sin\left(\frac{p}{1000^{\frac{2k}{d}}}\right), i = 2k \quad (3)$$

$$PE(p, i) = \cos\left(\frac{p}{1000^{\frac{2k}{d}}}\right), i = 2k+1 \quad (4)$$

where, $p$ is the position of the sequence, $d$ is encoding dimension and it equals to 32 in this study, $k \leqslant d/2$. This method not only provides a unique $PE$ for each position, but it also enables the relationship between two positions to be obtained via a straightforward linear function. By incorporating the positional encoding $PE$ into the differential entropy matrix $D$, a new matrix $D^*$ is obtained, which serves as the input for the self-attention layer (Equation 5).

$$D^* = D + PE \quad (5)$$

Consequently, the initial input $D$ now contains positional information.

The aim of the self-attention layer is to create a correlation within the input sequence and identify important information. To extract more comprehensive information from the differential entropy input, a multi-head self-attention mechanism is used. In our study, we have set the number of self-attention heads to eight. Each head can be expressed using a straightforward function:

$$SA = softmax(QK^T)V \quad (6)$$

where, $Q = D^* \cdot W_q$, $K = D^* \cdot W_k$, $V = D^* \cdot W_v$, $W_q$, $W_k$ and $W_v$ are learnable weight matrices. The eight head self-attention mechanism can be expressed as below:

$$M = Concat(SA_1, SA_2, \ldots SA_3)W_m \quad (7)$$

where, $W_m$ is a learnable weight matrix. $Concat$ represents the concatenation process utilized to merge the outputs of eight head self-attention layer.

The Feed-forward layer is composed of two linear functions with a ReLU activation function in between. The feed-forward layer takes the normalized value of the sum of $M$ and $D^*$.

$$I = M + D^* \quad (8)$$

$$I_n = \frac{I - E(I)}{\sqrt{Var(I)}} \quad (9)$$

$$O_{ff} = max(0, I_n W_1 + b_1)W_2 + b_2 \quad (10)$$

where, $O_{ff}$ represents the output of feed-forward layer, $E(I)$ is the expected value of $I$, $Var(I)$ is the standard deviation of $I$, $W_1$ and $W_2$ are learnable weight matrices, $b_1$ and $b_2$ are learnable biases.

The output of the complete self-attention module is the normalized sum of the matrices $D^*$ and $O_{ff}$.

$$O_{sum} = O_{ff} + D^* \quad (11)$$

$$O = \frac{O_{sum} - E(O_{sum})}{\sqrt{Var(O_{sum})}} \quad (12)$$

where, $O$ represents the output of the self-attention module, $E(O_{sum})$ is the expected value of $O_{sum}$, $Var(O_{sum})$ is the standard deviation of $O_{sum}$,

After obtaining the feature maps for each sample in the support subset $S_i$, we compute the final feature map $S_{fi}$ representing all $k$ support samples in the subset $S_i$ by taking the average of all feature maps. The feature map of the $j$-th sample in query set $Q$ is directly used as final feature map $x_{fj}^q$ gotten from Self-Attention Module.

### 3.3 Similarity Highlight Block

The similarity highlight block is employed to enhance the similarity component in both $S_{fi}$ and $x_{fj}^q$. This intensified feature map can aid the following driving-EEG determination block and drowsiness classification block in making better judgments.

This block comprises two similar cross-attention modules. Unlike the self-attention module where the number of input is only one, the cross-attention module takes two inputs: the Q feature and the K-V feature. The rest of the structure of the cross-attention module is similar to the self-attention module. This similar structure enables the cross-attention module to enhance the portion of the K-V feature that corresponds to the Q feature.

In our approach, we utilize each pair of feature maps $P_{fj}$ as inputs to the cross-attention modules. The first cross-attention module employs $S_{fi}$ as the Q feature and $x_{fj}^q$ as the K-V feature, while the second cross-attention module uses the opposite arrangement. By doing so, we can project the support feature onto the query space and the query feature onto the support space, resulting in an improved alignment of the similarity components of both $S_{fi}$ and $x_{fj}^q$.

Following each cross-attention module, a global average pooling operation is executed to derive the mean enhanced feature maps, denoted as $P_{fj}^* = \{S_{fi}^*, x_{fj}^{q*}\}_{i=1}^N$, over the 5 frequency bands.

### 3.4 Driving- EEG determination block

The driving-EEG determination block is a learning network that does anomalous signal detection, discriminating driving EEG samples. It has an advantage in that non-driving EEG samples do not need to be included as a class sample in the support set. We use the enhanced feature map $x_{fj}^{q*}$ as the input to this block. The block comprises a flatten layer, a ReLU activation, and a linear transformation.

The flatten layer transforms $x_{fj}^{q*}$ into a one-dimensional matrix $x_{ffj}^{q*}$. The ReLU activation and linear transformation can be represented by a simple mathematical expression:

$$DriEEG = max(0, x_{ffj}^{q*})W + b \quad (13)$$

Where, $W$ is learnable weight, $b$ is learnable weight, *DriEEG* is a vector with dimension of $2 \times 1$.

The classification of whether a query sample belongs to driving EEG samples is determined by locating the index with the highest value in the *DriEEG* vector.

### 3.5 Drowsiness Classification Block

The drowsiness classification block aims to assess the driver's level of drowsiness when the driving-EEG determination block identifies a query sample as a driving EEG sample. To achieve this goal, the Euclidean distance for the $i$-th class is computed between each pair in enhanced feature maps $P_{fj}^*$. The predicted class $y^q$ for the $j$-th sample in query set is determined by identifying the smallest distance value.

$$y_j^q = argmin(d_1, d_2, \cdots d_k) \quad (14)$$

Where, $d_i$ is the feature distance between the $S_{fi}^*$ and $x_{fj}^{q*}$, $(i = 1, \cdots, N)$.

## 4. Experimental Datasets and Implementation Details

### 4.1 Datasets

In this study, three datasets are employed: the SEED-VIG dataset, the Sustained-Attention Driving Task Dataset, and the SEED dataset. The SEED-VIG dataset and the Sustained-Attention Driving Task Dataset are utilized for analysing driver drowsiness, while the SEED dataset is applied for anomalous signal detection.

### 4.1.1 SEED-VIG

The data for this study was obtained from a simulated driving system, with the participation of 23 subjects. Each trial in the dataset comprises the EEG samples of a single subject throughout the entire 2-hour experiment. 17 channels' EEG samples are recorded from every subject and sampled at 200Hz. The dataset is labeled using PERCLOS, which is calculated based on eye-closing time recorded at 8-second intervals. A threshold of 0.7 is used to indicate drowsiness [22]. Each trial is divided into 885 samples, with each sample consisting of 8 seconds of EEG data.

$$PERCLOS = \frac{eye\ closing\ time}{interval\ time} \quad (15)$$

### 4.1.2 Dataset from Sustained-Attention Driving Task (SADT)

This dataset contains EEG data from 27 subjects who were asked to drive in a simulated system and maintain the virtual car's position at the center of the lane. The system generates random lane-departure events, causing the car to deviate from the original path. The counter-steering event recorded when the subject responds to the drift is used to determine the reaction time [23]. The drowsiness index can be calculated using Equation (17) with the react time $t$, and a value close to 1 indicates the subject is drowsy [31]. EEG samples were recorded from 32 channels for each subject, and a moving average filter with a window length of 10 was applied to smooth the drowsiness index. Each sample contains 3 seconds of EEG data, starting from the lane-departure event, and was sampled at a rate of 500Hz.

$$index = \max\left(0, \frac{1-e^{-(t-1)}}{1+e^{-(t-1)}}\right) \quad (17)$$

### 4.1.3 SEED

The data included in this dataset was not collected during a driving experiment. Instead, 15 subjects were asked to watch various movie clips designed to elicit positive or negative emotions. EEG data was recorded from 62 channels for each subject and used to analyse the changes in their emotional state caused by the movie clips [21]. In this study, we used the SEED dataset as a source of non-driving EEG data to evaluate the ability of the EEG-Fest algorithm to distinguish between driving and non-driving EEG data. The EEG data in this dataset was sampled at a rate of 200Hz.

To avoid abnormal results caused by artefacts from subjects' muscular or ocular motion present in EEG data, it is necessary to remove them. In this study, the Automatic Artifact Removal (AAR) plug-in for EEGLAB, a MATLAB toolbox, was utilized to perform this artifact removal process.

### 4.2 Implementation Details

The EEG-fest model is constructed under PyTorch deep learning platform. The model is trained and validated by a NVIDIA A100-80GB GPU with a batch size of 16 and 1, accordingly.

During training, the optimizer is Adam with beta1 = 0.5 and beta2 = 0.99. We set the initial learning rate to be 1e-5 and gradually increasing to 1e-4 over the course of 10 epochs. Loss functions are different between different tasks. For the regression task, we use the root mean square (RMSE) equation to compute the differences between the estimated results and the ground-truth. For the classification task, we use the cross-entropy function.

During evaluation, the model implementation is exactly the same with the training model without loss computation and parameters estimation.

## 5. Experiment Results and Discussions

### *5. 1 Vigilance Estimation Task*

The objective of the vigilance estimation task is to assess the effectiveness of self-attention mechanism for EEG-based signal processing. In this experiment, we utilize our self-attention based feature extraction block in conjunction with a fully connected layer to perform the vigilance regression, similar to previous studies conducted by other authors [30, 36, 22, 53].

#### 5.1.1 Training/Validation Data Split and Evaluation Metrics

We adhere strictly to the data split and experimental procedure as implemented by other authors [22, 30, 36, 53]. The task involves subject-specific regression for PERCLOS using the SEED-VIG dataset, with a total of 23 tests conducted, one for each subject. In each test, the 885 samples are divided into 5 sessions using 5-fold cross-validation.

For evaluation, the Pearson Correlation Coefficient (PCC) and Root Mean Squared Error (RMSE) are employed as evaluation metrics for this task [54]. PCC provides the linear relationship between prediction and ground truth, taking on a value within the range of [-1, 1]. A PCC value closer to 1 indicates a strong linear relationship between the prediction and ground truth. RMSE gives us the difference between prediction and ground truth. Smaller RMSE shows the higher similarity. Equations for PCC and RMSE are listed below:

$$PCC(T,P) = \frac{\sum_{i=1}^{n}(t_i-\bar{t})(p_i-\bar{p})}{\sqrt{\sum_{i=1}^{n}(t_i-\bar{t})^2 \sum_{i=1}^{n}(p_i-\bar{p})^2}} \quad (18)$$

$$RMSE(T,P) = \sqrt{\frac{1}{n}\sum_{i=1}^{n}(p_i-t_i)^2} \quad (19)$$

Where, $t$ is the ground truth value; $\bar{t}$ is the mean of all ground truth values, $p$ is the prediction value, $\bar{p}$ is the mean of all prediction values, $n$ is the total number of prediction value and ground truth.

#### 5.1.2 Results and Analysis

Table 2 presents a comparison between our proposed self-attention-based method (SAM) and other existing models, with the average values and standard deviations calculated for the 23 tests. Our model achieves a PCC of 0.96±0.01, which is the highest among all models, and 17% higher than the second highest result. As for the RMSE results, the proposed SAM is 0.03±0.01, which is smaller than all other models, and 67% smaller than the second highest results. These outcomes suggest that our SAM is more effective in extracting features from EEG samples and delivering more accurate predictions. Thus, the proposed SAM has great potential for feature extraction under few-shot learning.

Additionally, Figure 3 illustrates the PERCLOS curve for four subjects, displaying the predicted values in comparison to the ground truth and visually confirming the strong performance of SAM. According to Figure 3, the predicted PERCLOS pattern reconstruct the ground truth pattern only with a few noises. In the real-world implementation, these occasional impulse noises can be easily filtered through a bandwidth filter.

Table 2 Result's Comparison

| Model | PCC(↑) | RMSE(↓) |
|---|---|---|
| GELM[30] | 0.70±0.10 | 0.10±0.03 |
| LSTM[36] | 0.82±0.08 | 0.09±0.03 |
| SVR[22] | 0.70±0.23 | 0.13±0.03 |
| DNNSN[53] | 0.72±0.17 | 0.12±0.04 |
| **SAM** | **0.96±0.01** | **0.03±0.01** |

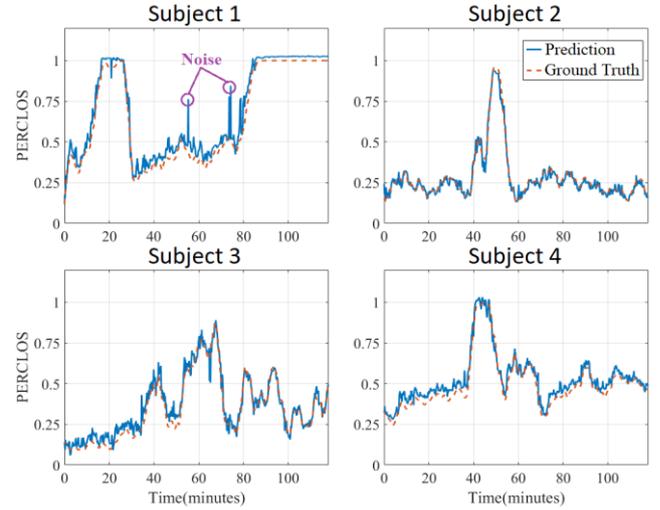

Figure 3: Comparison of PERCLOS prediction and Ground Truth of four subjects

### *5.2 Few-shot Learning Results*

### *5.2.1 Dataset Selection*

In the few-shot learning studies, we employ two driving-related datasets, which is the SEED-VIG and the SADT dataset. Since the EEG-Fest contains a non-driving EEG determination module to detect anomalous signals, a non-driving EEG dataset, SEED, is used as anomalous samples. During the experiment, the dataset setup is as follows: The first setup comprises of SEED-VIG as the driving EEG dataset and SEED as the non-driving EEG dataset. The second setup involves SADT as the driving EEG dataset and SEED as the non-driving EEG dataset. We treat two dataset setups independently and train two independent EEG-Fest models based on these setups.

For the first set of datasets (SEED-VIG+SEED), we choose data samples of 16 subjects from SEED-VIG and 10 subjects from SEED for model training, and data samples of another 7 subjects from SEED-VIG and 5 subjects from SEED for validation. In addition, we selected 17 same EEG channels data from both SEED-VIG and SEED (Figure 4a). All samples from SEED dataset are labeled as non-driving EEG, while samples from SEED-VIG dataset are labeled as non-drowsy or drowsy based on their PERCLOS values.

For the second set (SADT+SEED), we choose data samples of 19 subjects from SADT and 10 subjects from SEED for model training, and data samples of another 8 subjects from SADT and 5 subjects from SEED for validation. We used data from 26 same channels from both SADT and SEED (Figure 4b). Since the SEED dataset EEG signals' sampling frequency is 200 Hz, which is smaller than the SADT (500 Hz), we upsample the SEED dataset through linear interpolation to match the sample rate of SADT. Similar to the first setup, the SEED dataset samples are labeled as non-driving EEG, whereas the samples from the SADT dataset are labeled as either non-drowsy or drowsy based on their drowsiness index.

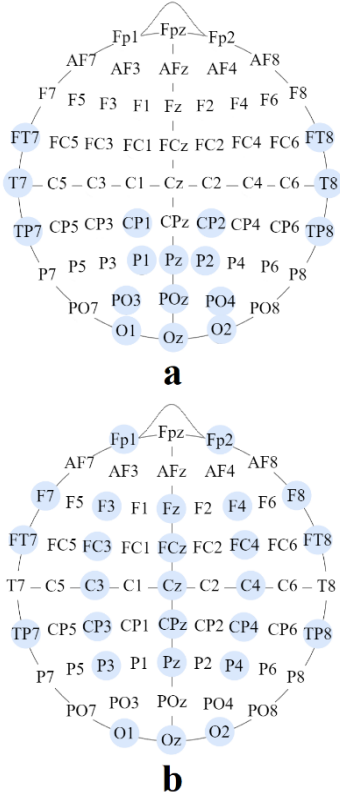

Figure 4: Channels in blue are the selected EEG Channels in International 10–20 system for two sets' experiments.

### 5.2.2 Training/Validation Data Split and Evaluation Metrics

Both training and the validation dataset requires support samples and query samples. In this study, we set the number of support values to $K = 1, 5, 10,$ and $20$. The number of total samples in training and validation sets is shown in Table 3. In both the training and validation sets, 40 support samples are selected at random from each subject to create a support set for the 20-shot setting, while the remaining driving-EEG samples and non-driving EEG samples are used to form the query set. The support sets for 1-shot, 5-shot, and 10-shot are all subsets of the support set for 20-shot. Additionally, we constructed two comparison modules, based on Convolutional Neural Network (CNN) [60] and Long Short-Term Memory (LSTM) [36], as a baseline comparison for the EEG-Fest. The CNN and LSTM models were tested using the entire training set for initial training, followed by fine-tuning with all support samples in support set for 20-shot from the evaluation set. This was done to assess the performance of traditional supervised learning-based EEG algorithms in the same cross-subject setting.

Table 3 Data Samples for Few-Shot Experiment

|  |  | SEED-VIG +SEED | SADT +SEED |
|---|---|---|---|
| Training | Query | 10408 | 10314 |
|  | Support | $2K$ | $2K$ |
| Validation | Query | 4460 | 4582 |
|  | Support | $2K$ | $2K$ |

To evaluate the performance of EEG-Fest, we use the accuracy for each class and the average F1 score across all classes as the evaluation metrics. The accuracy is simply the ratio between the correct estimated samples and the total samples for each class. For the F1 score, the equation is:

$$F1\ score = \frac{TP}{TP+\frac{1}{2}(FP+FN)} \quad (20)$$

where $TP$, $FP$, and $FN$ are the number of true positive, false positive, and false negative results. A higher F1 score signifies higher precision and lower recall for a specific class.

We repeat the experiment for five times with five different randomly selected groups of support samples.

### 5.2.3 Results and Analysis

The evaluation results are illustrated in Table 4 and Table 5. As the EEG-Fest is the first few-shot learning algorithm for driver drowsiness, there are no other few-shot approaches available for comparison. Therefore, we compare our few-shot results with two traditional supervised learning results (CNN & LSTM). According to Table 4 and Table 5, our proposed model outperforms traditional supervised learning-based EEG algorithms in a cross-subject setting, despite the fact that [60] and [36] reported high performance when using sufficient subject data during training. Prior to utilizing the training set to train the entire EEG-Fest, we begin by using all the training samples that have a known ground truth to train the feature extraction block. The final trained parameters of this block are then utilized as the initial values for EEG-Fest's feature extraction block, following the approach used by [42] in their research. As per [42], utilizing the pretrained parameters can enhance the final classification outcome. Our experimental results validate this claim, as even in a 1-shot setting, our classification accuracy is higher than the baseline.

Overall, the performance of our approach improves with an increase in the number of shots. In the 1-shot setting, our approach results in an F1 score improvement of 0.2 and 0.09 on the first set of datasets and 0.12 and 0.02 on the second set of datasets. In the final 20-shot setting, our approach achieves a better F1 score, specifically 0.86 on the first set of datasets and 0.83 on the second set of datasets. Moreover, our approach exhibits better classification ability on all three classes.

Our further observations indicate that both the LSTM-based approach and attention-based approach are more effective in detecting anomalous samples (non-driving EEG) than the CNN-based approach. This is likely due to the fact that the LSTM-based approach and attention-based approach can capture the long sequence of EEG samples, leading us to conclude that an approach capable of extracting relatively long sequential features from raw data is better suited to anomaly detection. However, our attention-based approach performs even better than the LSTM-based approach, with mean accuracies of 0.91 and 0.9 for non-driving samples in the final 20-shot setting on two sets of datasets, respectively. These accuracies are better than those achieved by the LSTM-based approach by 8% and 10%.

Table 4 Result from First Set of Datasets

| Method   | k-shot | Driving    |        | Non-Driving | F1 Score |
|----------|--------|------------|--------|-------------|----------|
|          |        | Non-Drowsy | Drowsy |             |          |
| CNN      | -      | 0.49       | 0.62   | 0.61        | 0.58     |
| LSTM     | -      | 0.54       | 0.67   | 0.83        | 0.69     |
| EEG-Fest | 1      | 0.70       | 0.78   | 0.86        | 0.78     |
|          | 5      | 0.74       | 0.80   | 0.87        | 0.81     |
|          | 10     | 0.76       | 0.84   | 0.89        | 0.83     |
|          | 20     | **0.78**   | **0.86** | **0.91**  | **0.86** |

Table 5 Result from Second Set of Datasets

| Method   | k-shot | Driving    |        | Non-Driving | F1 Score |
|----------|--------|------------|--------|-------------|----------|
|          |        | Non-Drowsy | Drowsy |             |          |
| CNN      | -      | 0.58       | 0.73   | 0.51        | 0.65     |
| LSTM     | -      | 0.64       | 0.77   | 0.80        | 0.75     |
| EEG-Fest | 1      | 0.66       | 0.79   | 0.87        | 0.77     |
|          | 5      | 0.68       | 0.80   | 0.88        | 0.79     |
|          | 10     | 0.72       | 0.82   | 0.89        | 0.81     |
|          | 20     | **0.74**   | **0.84** | **0.90**  | **0.83** |

### 5.3 Ablation Study

For our ablation study, we selected the first set of datasets and utilized the same dataset settings as described in section 5.2.2.

#### 5.3.1 Influence of Different Similarity Measurement

Table 6 Mean Accuracy of non-drowsy class based on different similarity measurements

| k-shot | Euclidean | Standard Euclidean | Cosine | Correlation |
|--------|-----------|--------------------|--------|-------------|
| 1      | **0.70**  | 0.69               | 0.68   | 0.68        |
| 5      | **0.74**  | 0.73               | 0.71   | 0.71        |
| 10     | **0.76**  | 0.73               | 0.74   | 0.74        |
| 20     | **0.78**  | 0.78               | 0.75   | 0.75        |

Table 7 Mean Accuracy of drowsy class based on different similarity measurements

| k-shot | Euclidean | Standard Euclidean | Cosine | Correlation |
|--------|-----------|--------------------|--------|-------------|
| 1      | **0.78**  | 0.76               | 0.77   | 0.77        |
| 5      | **0.80**  | 0.78               | 0.79   | **0.80**    |
| 10     | **0.84**  | 0.80               | 0.83   | 0.83        |
| 20     | **0.86**  | 0.85               | 0.85   | **0.86**    |

The use of different similarity measurements may lead to variations in the results. To assess the impact of different similarity measurements on the classification of non-drowsy and drowsy states in driving-EEG samples, we used four different similarity measures to calculate the distance between each pair of entries in the enhanced feature maps $P_{fj}^*$. The four different measurement we select are: Euclidean distance,

Standard Euclidean distance, Cosine distance and Correlation distance. We report the mean accuracy of repeated five experiment in the Table 6 and Table 7.

Both tables show that Euclidean distance is more appropriate in our approach. The results indicate that the largest difference between Euclidean distance and the other three distances is 3% for the non-drowsy class and 4% for the drowsy class.

*5.3.2. Influence of Driving-EEG Determination Block*

We remove this block in our proposed EEG-Fest. Additionally, we have included non-driving EEG samples in the support set since the original version of EEG-Fest did not require non-driving EEG samples in the support set for the driving-EEG Determination block. To identify whether an EEG sample is a non-driving sample, we calculate the distance between each pair of entries in the enhanced feature maps $P_{fj}^*$, similar to how we classify non-drowsy and drowsy samples.

The result of mean accuracy of 5 repeated experiment is shown in Table 8. With the help of the driving-EEG determination block, our approach exhibits a better performance on detection non-driving samples with the improvement of 25%, 24%, 22%, 20% for 1-shot, 5-shot, 10-shot and 20-shot evaluation, respectively. Such improvement demonstrates the importance of Deriving-EEG Determination block.

Table 8 Mean Accuracy of Non-Driving samples from the approach with driving-EEG determination block and without driving-EEG determination block

| $k$-shot | With | Without |
|---|---|---|
| 1 | **0.86** | 0.61 |
| 5 | **0.87** | 0.63 |
| 10 | **0.89** | 0.67 |
| 20 | **0.91** | 0.71 |

*5.3.3 Influence of Similarity Highlight Block*

In this test, we have omitted the similarity highlight block because we want to examine its impact on the classification performance of drowsy and non-drowsy samples. Instead, we directly calculate the distance between the feature maps in $P_{fj} = \{S_{fi}, x_{fj}^q\}_{i=1}^N$ for each class that are produced by the feature extraction block. The drowsiness class is determined based on the smallest distance value.

According to Table 9, the inclusion of the similarity highlight block results in an increase in mean accuracy for non-drowsy samples by 3%, 6%, 4%, and 4% in the 1-shot, 5-shot, 10-shot, and 20-shot evaluation setting, respectively. In addition, the block also leads to an improvement in the performance of drowsy samples by 10%, 9%, 9%, and 8% in the same setting, respectively. The effect of similarity highlight block is shown by the observed improvement.

Table 9 Mean Accuracy of drowsy and non-drowsy samples from the approach with similarity highlight block and without similarity highlight block

| $k$-shot | With | | Without | |
|---|---|---|---|---|
| | Non-Drowsy | Drowsy | Non-Drowsy | Drowsy |
| 1 | **0.70** | **0.78** | 0.67 | 0.68 |
| 5 | **0.74** | **0.80** | 0.68 | 0.71 |
| 10 | **0.76** | **0.84** | 0.72 | 0.75 |
| 20 | **0.78** | **0.86** | 0.74 | 0.78 |

## 6. Conclusion

In this work, we propose a few-shot model (EEG-Fest), aiming to detect the driver's drowsiness as well as anomalous signal with limited sample data.

We first introduce a self-attention-based method (SAM) as a feature extraction tool in our EEG-Fest model. To demonstrate SAM's practicality and effectiveness, we used it to predict the PERCLOS of 23 subjects. The results indicate a Pearson Correlation Coefficient of 0.96±0.01 and a Root Mean Squared Error of 0.03±0.01, signifying that our model is well-suited to serving as a feature extraction tool for EEG data samples.

Afterward, we tested the effectiveness of our proposed EEG-Fest in a cross-subject setting using two sets of datasets. The first set comprises the SEED-VIG dataset and the SEED dataset, while the second set comprises the SADT dataset and the SEED dataset. As a baseline comparison, we also conducted the same evaluation on a CNN-based model and an LSTM-based model. The results demonstrate that our proposed approach outperforms the other two conventional algorithms for drowsiness detection as well as anomaly detection.

Lastly, we conducted an ablation study from three perspectives to assess the impact of different similarity measurements, the driving-EEG determination block, and the similarity highlight block. Our findings indicate that the Euclidean distance is the most effective similarity measurement method in our approach. The driving-EEG determination block enhances the detection of non-driving samples, while the similarity highlight block improves the accuracy of drowsiness prediction in drivers.

Although this study demonstrates the applicability of the attention mechanism and few-shot approach in a cross-subject setting, further validation is necessary in our future work. Our model was only tested on two driving EEG datasets and one non-driving EEG dataset. Therefore, in future research, we need to consider more datasets to evaluate the full potential of our proposed model.

## References

[1] NHTSA. Drowsy driving, 2020